\documentclass[10pt,twocolumn,letterpaper]{article}

\usepackage[pagenumbers]{wacv} 

\usepackage[numbers]{natbib}
\usepackage{tabularx}
\usepackage{siunitx}
\usepackage{booktabs}
\usepackage{multirow}
\usepackage{makecell}
\usepackage{array}

\newcommand{\showornot}[1]{{\color{red}{}}} 

\definecolor{wacvblue}{rgb}{0.21,0.49,0.74}
\usepackage[pagebackref,breaklinks,colorlinks,allcolors=wacvblue]{hyperref}

\def\wacvPaperID{380} 
\def\confName{WACV}
\def\confYear{2026}

\title{Posture-Driven Action Intent Inference for Playing style and Fatigue Assessment}

\author{Abhishek Jaiswal\\
Indian Institute of Technology Kanpur\\
India\\
{\tt\small abhi.jaiswal44@gmail.com}
\and
Nisheeth Srivastava\\
Indian Institute of Technology Kanpur\\
India\\
{\tt\small nsrivast@iitk.ac.in}
}

\begin{document}
\maketitle

\begin{abstract}

Posture-based mental state inference has significant potential in diagnosing fatigue, preventing injury, and enhancing performance across various domains. Such tools must be research-validated with large datasets before being translated into practice. Unfortunately, such vision diagnosis faces serious challenges due to the sensitivity of human subject data.
To address this, we identify sports settings as a viable alternative for accumulating data from human subjects experiencing diverse emotional states. We test our hypothesis in the game of cricket and present a posture-based solution to identify human intent from activity videos.
Our method achieves over 75\% F1 score and over 80\% AUC-ROC in discriminating aggressive and defensive shot intent through motion analysis. 
These findings indicate that posture leaks out strong signals for intent inference, even with inherent noise in the data pipeline. Furthermore, we utilize existing data statistics as a weak supervision to validate our findings, offering a potential solution for overcoming data labelling limitations.
This research contributes to generalizable techniques for sports analytics and also opens possibilities for applying human behavior analysis across various fields.

\end{abstract}
    
\section{Introduction}
\label{sec:intro}

\begin{figure*}[bthp]
    \centering
        \includegraphics[width=.80\linewidth]{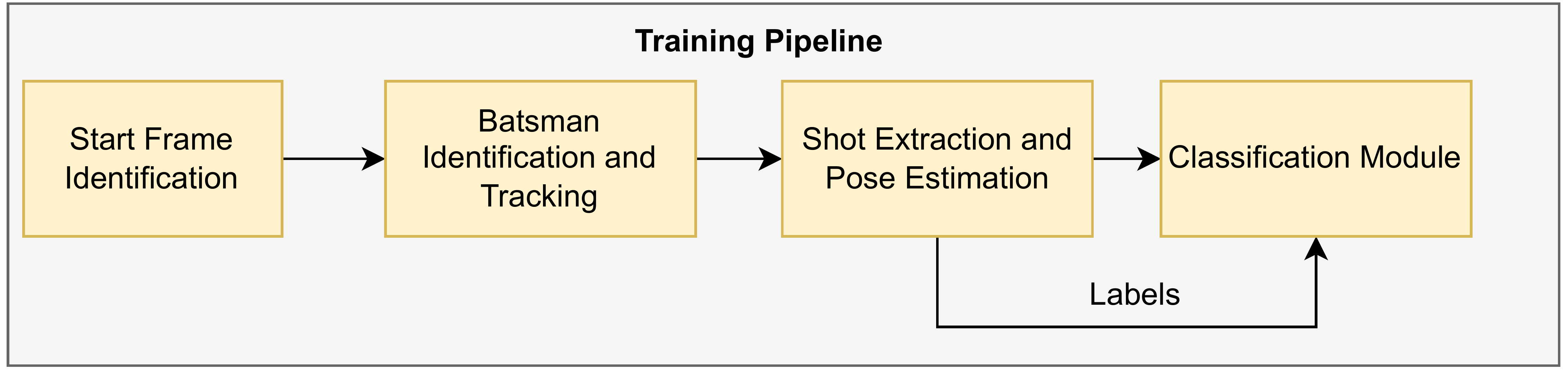}
    \caption{ Training Pipeline for Intent Classification}
    \label{fig:crik_motion}
\end{figure*}

Pose and motion are established biomechanical indicators in clinical practice, aiding in the diagnosis and treatment of health conditions. Research has shown that upright posture can improve mood and reduce fatigue levels~\cite{nair2015slumped,wilkes2017upright}.
Reciprocally, stress and fatigue can impair muscle control and, in turn, negatively affect posture~\cite{paillard2012effects, lin2009acute, schieppati2003neck}. This bidirectional relationship between posture and mental states presents an opportunity to develop new evaluation and assessment tools, particularly in physical training, where unmanaged fatigue could significantly increase the risk of injury~\cite{slobounov2008fatigue, mclean2009fatigue, schampheleer2024mental}.
Physical activity, especially sports training, could thus benefit from techniques that can indirectly infer player fatigue. Therefore, technical exploration is warranted for such automatic assessment tools to improve training regimens and reduce injuries.

Posture-based activity identification has already been explored in Human Action Recognition (HAR) studies~\cite{roitberg2021let,wang2013approach,martin2019drive,singh2019multi,nadeem2021automatic,10.1145/3604915.3608816}. However, the association between mental state, intent, and posture has not been adequately researched in the vision domain. Studies in health and biomechanics have already established these links between mental states and posture. For example, \citet{rosario2016angry} found the correlation between anger and shoulder elevation and hyperextension of the knees. Similarly, depression has been shown to visibly affect posture~\citep{canales2010posture,dehcheshmeh2024correlation}. Yet, these results have not translated into vision-based tools due to the scarcity of sensitive, labeled health datasets, limiting the exploration of these connections fully.

To address this gap, we explore vision-based detection of action intent from posture. Identifying sports as a compelling application domain for intent-driven actions, we pose the problem as that of posture-based intent inference from sports data clips. Sports offer a rich environment where athletes perform actions under varying mental states, and they are often associated with match statistics, which can serve as weak supervisory signals for labeling actions.

Specifically, we analyze the game of cricket, which, similar to baseball, is played between two teams with a batter hitting the ball (called a batter's shot) thrown by a bowler. We investigate how well machine learning models can classify batters' shots into aggressive and defensive intents using posture and motion data. Such analysis can provide insights into a player's playing style and alert support staff if a deviation from a player's natural style might signal an underlying issue.
Additionally, we also explore how publicly available match statistics can support the analysis of mental state inference.

The ability to infer mental state from posture has broad applications. In healthcare, it can guide immediate treatment plans depending on patient's anxiety and fatigue. In sports, real-time intent inference can alert coaches or medical staff when an athlete is at risk of exhaustion or injury. More broadly, non-invasive biomechanical monitoring can enhance support systems in high-pressure environments, contributing to safer and more responsive human assistive technologies.


To summarize, the contributions of this work are as follows:
\begin{itemize}
    \item We release a pose csv dataset for cricket shots played under aggressive and defensive mental states (Section \ref{dataset}).
    \item  We develop a generalizable pipeline to estimate shot intent from motion video clips in cricket (Section \ref{experiments} and \ref{results}).
    \item Using match statistics based analysis , we explore the validity of our results and explore applications to support player assessment (Section \ref{case_study}).

\end{itemize}






\begin{figure*}[bt]
  \centering
  \begin{minipage}[t]{0.48\textwidth}
    \vspace{0pt}
    \centering
    \includegraphics[width=0.9\linewidth]{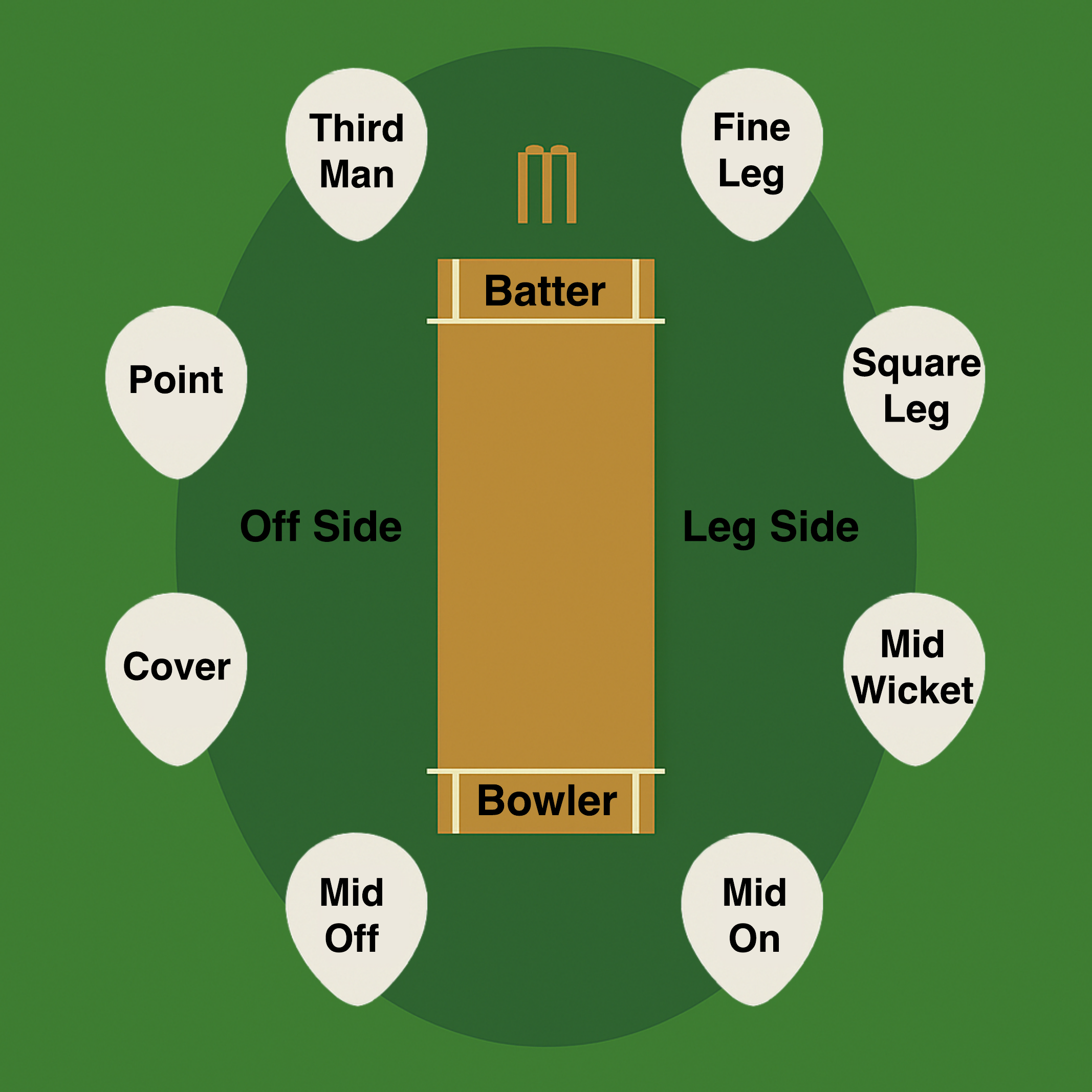}
    \caption{Cricket Field—simplistic schematic area representation.}
    \label{fig:cric_field}
  \end{minipage}%
  \hfill
  \begin{minipage}[t]{0.48\textwidth}
    \vspace{0pt}
    \centering
    \includegraphics[width=\linewidth]{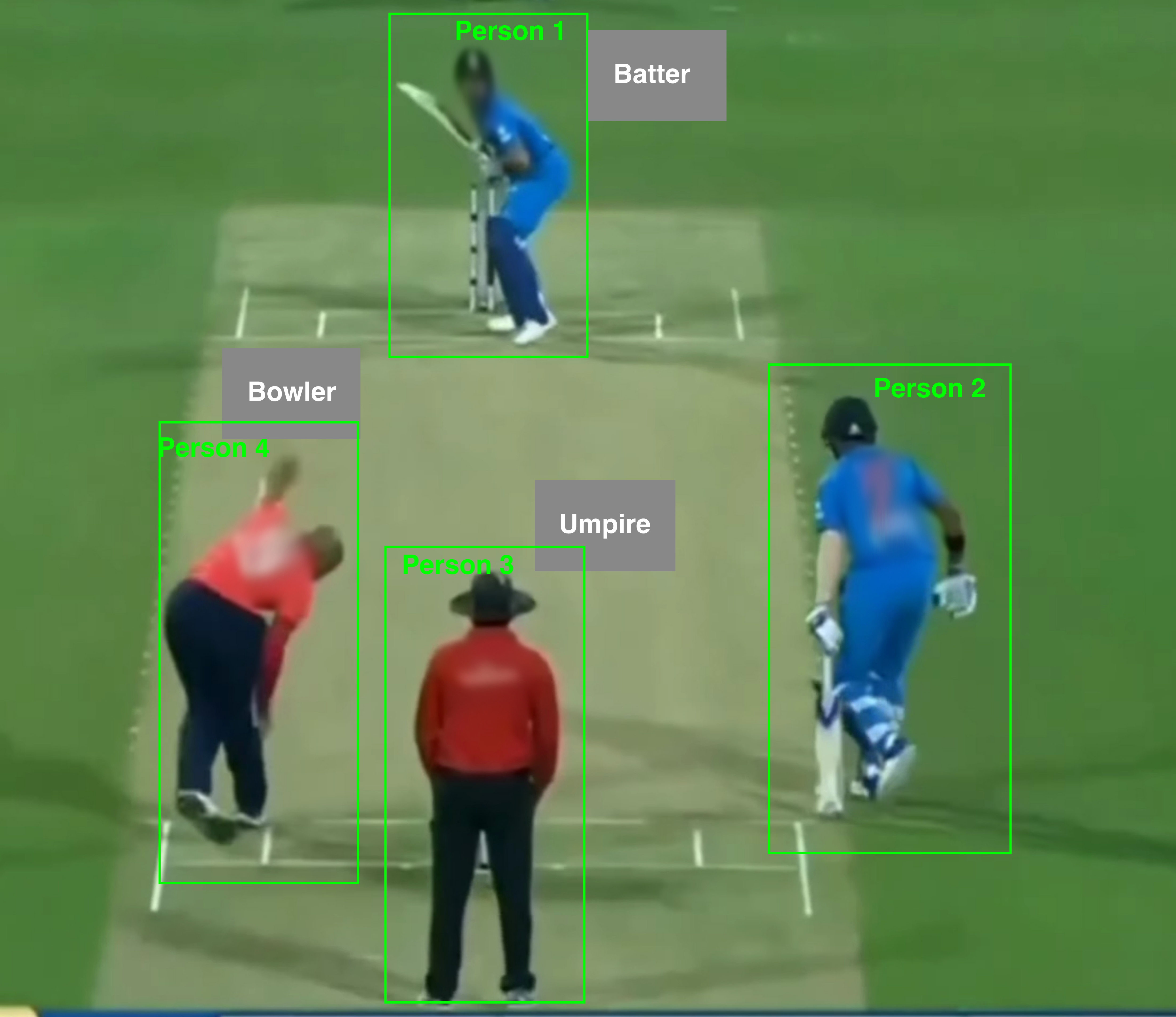}
    \caption{Start frame of an input sequence marking the batter ready to play shot~\citep{indvivek2025youtube}.}
    \label{fig:start_frame}
  \end{minipage}
\end{figure*}

\section{Intent Inference: Application Domains}
\label{related_works}
The background of intent analysis from visual cues draws inspiration from multiple related fields. Broadly, intent inference can be considered a subset of action recognition. However, in typical action recognition, the gestures or actions to be recognized remain fixed, like in walking and eating.  In contrast, intent inference is more complex, as similar intents may manifest through different actions. 

The premise of our work is that posture conveys subtle signals about mental states and should be further explored to develop assistive technologies. Relevant studies to this theme,  especially in sports, have applied visual analytics to understand team tactics or forecast sports actions~\cite{lord2020methods,wunderlich2021forecasting}. Tactics analysis has been explored in various sports such as tennis, football, and volleyball~\cite{wu2021tacticflow,goes2021unlocking,niu2012tactic,kong2022spatio}. For sports action forecasting,~\citet{felsen2017will} provide a generic framework to anticipate next moves in water polo and basketball directly from visual inputs. Both kind of studies utilize posture-based inference and incorporate elements of intent-based analysis in different forms.

Other related applications have analyzed pedestrian intent at crosswalks to assist autonomous driving and improve road user safety~\cite{zhao2021action,sharma2022pedestrian}. For instance, ~\citet{liu2020spatiotemporal} predict pedestrian intent for future street crossing using graph convolutions to model pedestrians' spatiotemporal context. Such works apply intent inference to examine and predict human behavior.

In the healthcare context, posture has been shown to have correlations with mental states such as anger, anxiety, and depression~\cite{feldman2020gait,rosario2016angry}. This association between physical and mental states could be explored further for potential visual diagnostic assistance.

As a promising application, analyzing mental intent can help identify highly energy dissipating aggressive actions, enabling more accurate tracking of athlete fitness during physical activity. Related to this theme, \citet{kooij2016multi}  have explored the identification of aggressive motion for safety surveillance.
 Energy expenditure from physical activity can also be tracked using sensor systems, as demonstrated by~\citet{sazonova2014posture}.
These use cases highlight the need for further technological innovations to enhance visual diagnostic tools.

In this work, we address the problem of identifying intent from posture and envisage potential applications for such tools. Taking sports as a potential avenue for such inference, we focus on the game of cricket.  International sports, such as cricket, generate vast amounts of data over digital media, which can be leveraged for sports analytics~\cite{bertasius2017baller,felsen2017will}. In such broadcast sports, the game statistics can also provide weak supervision, serving as labels for game actions. We further explore this approach to validate our results in intent inference.

Readers are encouraged to refer to supplementary file Section 3 for definitions of common cricket terms, which may enhance the understanding this work.

\section{Cricket Shot Intent Dataset (CSID): Design and Composition}
\label{dataset}
In physical activity, it is typical of athletes to expend more energy when acting with aggressive intent to achieve their goals.
Following this rationale, and for simplicity, we use the terms energy and aggressiveness interchangeably in this study, with high-energy shots representing aggressive intent, and low-energy shots indicating defensive intent. 
To ensure consistent analysis and labeling of the video clips, we infer the shot energy and aggressiveness through visual inspection of the batter’s shot speed.

We built our dataset by extracting clips of batters' shots from YouTube cricket match videos, maintaining a separate data folder for each match.
 Clips that did not capture a complete shot or contained too much extra footage were excluded during initial filtering. All remaining clips were manually annotated as either high- or low-energy shots. Only examples from the extremes of the energy spectrum were included; ambiguous, intermediate-energy shots were intentionally omitted to sharpen class separation. To maintain sufficient shot count, matches containing a small number of usable shots were merged into a single folder. Additionally, exclusive videos featuring shots from the same batter were also combined.

This process resulted in a curated dataset across eleven data folders, comprising over 2,500 shot clips labeled as high- or low-energy (\Cref{tab:csid}). Annotations underwent random verification by an independent annotator to ensure labeling consistency. The dataset includes clips from all three major international cricket formats: One Day Internationals (ODIs), Twenty20 (T20), and Test matches. Finally, we use Google's Mediapipe Pose framework~\cite{lugaresi2019mediapipe} to extract pose sequences from these clips and these sequences were used to train our classifiers.

For posture samples of high- and low-energy shots, please refer to CSID-Vizualisations folder in the supplementary material.


\section{Automated Shot Segmentation and Sequential Modelling}
\label{experiments}
In sports settings, intent inference could be conducted at both the player and team levels, with players fulfilling distinct roles within a team. In this work, we focus on intent analysis for the batter who strikes the ball. To achieve this, we extract the relevant segment of the batter's shot from the match video. By applying pose estimation to these extracted clips, we obtain motion data for the batter's body joints which we use for intent classification (see \Cref{fig:crik_motion} for overall training pipeline).

\showornot{
\textbf{HOW TO IDENTIFY WHICH FEATURE TO USE FOR ALL THESE MODELS?}
Using these extracted poses along with the video based shot labels we train multiple classifiers for shot intent classification.
}
\subsection{Player Shot Extraction Pipeline}
The initial steps involve identifying the batter as he prepares to play a shot and extracting the corresponding segment from the match video. For this, we employ the YOLO~\cite{yolo11_ultralytics} person detector. This way, we get the locations of all individuals within each frame. Next, we apply heuristics to determine whether any detected individual's position corresponds to the typical location of the batter at the moment a shot is about to be played (Figure \ref{fig:start_frame}).

After pinpointing the first frame in which the shot occurs, we track the batter as long as he remains within a predefined, fixed-width region of the screen. When the tracked batter exits this region, we infer that the shot has been completed and record the temporal pose data to this point as the duration of the clip. This approach leverages the fact that, after the ball is hit, the camera follows the ball's trajectory, causing the batter to move out of the frame.




\begin{figure*}[tbhp]
  \centering
  \begin{minipage}[b]{0.60\textwidth}
    \vspace{0pt}
    \centering
    \includegraphics[width=\linewidth]{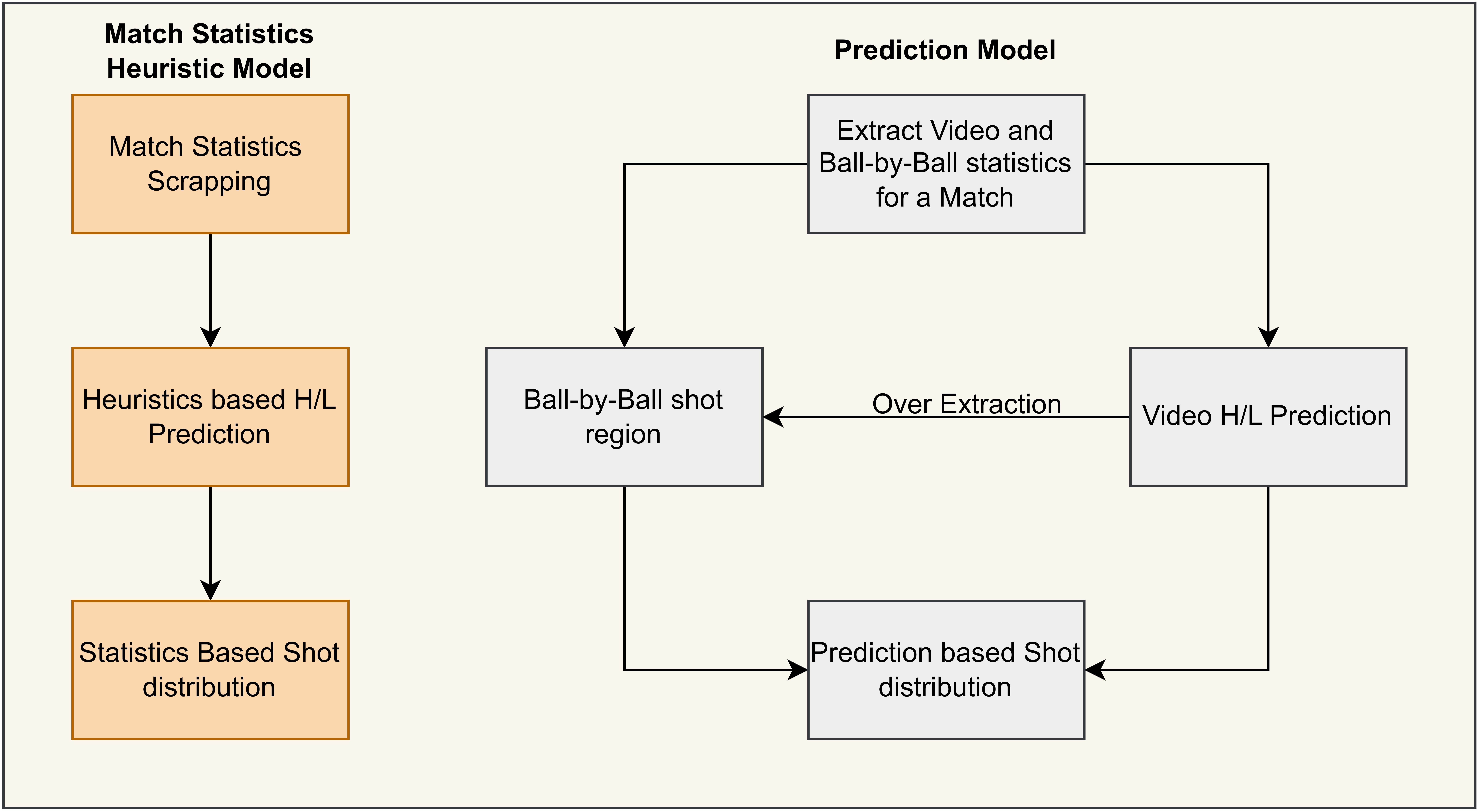}
    \caption{Workflow of the statistics-based heuristic model and vision-based prediction model for analyzing the batter's shot region distribution.}
    \label{fig:commentary_pipeline}
  \end{minipage}%
  \hfill
  \begin{minipage}[b]{0.38\textwidth}
    \vspace{0pt}
    \centering
    \renewcommand{\arraystretch}{1.2}
    \begin{tabular}{@{}lrr@{}}
      \toprule
      \textbf{Folder} & \textbf{High} & \textbf{Low} \\
      \midrule
      Folder 1        & 120   & 52    \\
      Folder 2        & 75    & 12    \\
      Folder 3        & 59    & 28    \\
      Folder 4        & 187   & 44    \\
      Folder 5        & 29    & 102   \\
      Folder 6        & 17    & 108   \\
      Folder 7        & 105   & 81    \\
      Folder 8        & 452   & 733   \\
      Folder 9        & 87    & 80    \\
      Folder 10       & 86    & 65    \\
      Folder 11       & 18    & 71    \\
      \midrule
      \textbf{Total}  & \textbf{1235} & \textbf{1376} \\
      \bottomrule
    \end{tabular}
    \captionof{table}{High and Low File Counts for Each Folder.}
    \label{tab:csid}
  \end{minipage}
\end{figure*}

\subsection{Classification Models}
We explore several time series classification models on our dataset for mental state inference of the shot played.

\textbf{1D Convolutional Neural Network (1D CNN)}: Processes multivariate time series using one-dimensional convolutional layers to extract local temporal patterns. The convolution filter slides along the time axis, processing all input features within the kernel window at each step.

\textbf{Long Short-Term Memory (LSTM)}: Utilizes gating mechanisms to retain or forget information, making it suitable for capturing long-range temporal dependencies in sequential data.

\textbf{LSTM Autoencoder}:  Employs an encoder-decoder architecture to learn low-dimensional representations of sequential input. Unlike standard autoencoders, this model is jointly trained with a classifier using a combined loss function that includes both reconstruction and classification errors, enabling it to learn features useful for both data reconstruction and activity classification.

\textbf{Motion Range Model}: Calculates the motion range (max-min) for each feature across the time series. The resultant feature vector is used to train a random forest classifier for high/low energy action recognition. This approach captures movement variability, often linked to high-energy shots in cricket.

\textbf{Two-Stream Adaptive Graph Convolutional Network (2s-AGCN) \cite{shi2019two}}: Based on Spatio-Temporal Graph Convolutional models (STGCN) ~\cite{yan2018spatial}, the 2s-AGCN model adaptively learns graph topologies for action recognition by processing first-order features (joint positions) and second-order features (bone vectors). It employs a two-stream architecture, where one stream captures the dynamics of joint positions and the other models bone information, allowing the network to effectively learn spatial and temporal dependencies in human skeletal data for improved action recognition performance.

All the models use shoulder, elbow, wrist, hip, knee, ankle, and heel joint coordinates as inputs, ensuring consistency in feature representation. The 2s-AGCN additionally uses the nose joint and incorporates joint prediction confidence values as part of its architecture. All models are trained with early stopping on the F1 score for up to 2500 epochs. Each time series feature has its initial ten values removed and capped to a maximum length of fifty.


\section{Model Performance on Energy Inference}
\label{results}

In this section, we compare different models for intent classification and investigate how classification performance varies with changes in the clip duration of the batter's shot.

\begin{table*}[bthp]
\centering

\vspace{0.5em}

\begin{tabular}{lccc}
\hline
\textbf{Classifier} & \textbf{Accuracy} & \textbf{AUC-ROC} & \textbf{F1 Score} \\
\hline

LSTM & 0.75 $\pm$ 0.10 [0.73, 0.76] & 0.81 $\pm$ 0.08 [0.80, 0.83] & 0.71 $\pm$ 0.14 [0.68, 0.73] \\
Motion Range* & 0.73 $\pm$ 0.10 [0.66, 0.80] & 0.79 $\pm$ 0.09 [0.73, 0.85] & 0.70 $\pm$ 0.12 [0.62, 0.78] \\

LSTM AE & 0.73 $\pm$ 0.14 [0.70, 0.76] & 0.79 $\pm$ 0.13 [0.77, 0.81] & 0.72 $\pm$ 0.18 [0.69, 0.75] \\
2s-AGCN & $\mathbf{0.78 \pm 0.10 [0.76, 0.80]}$ & $\mathbf{ 0.87 \pm 0.06 [0.86, 0.88]}$ & $\mathbf{0.78 \pm 0.12 [0.75, 0.80]}$ \\

1D CNN & $\mathbf{0.77 \pm 0.07 {[}0.75, 0.78{]}}$ & 0.83 $\pm$ 0.07 {[}0.82, 0.85{]} & $\mathbf{0.77 \pm 0.08 {[}0.76, 0.78{]}}$ \\

\hline

\end{tabular}

\caption{Performance for various models showing Mean $\pm$ standard deviation and 95\% confidence intervals.
*Motion range model does not need a validation set so leave-one-out-cross-validation results compared. Total Dataset Size: High clips: 1236, Low clips: 1376.
}
\label{tab:model_performance}
\end{table*}

\subsection{Performance on Intent Classification}
Our evaluation was conducted using ordered leave-pair-out cross-validation on data from eleven folders. For each iteration, one folder becomes the validation set, one folder becomes the test set, and the remaining nine folders are used for training, creating $^{11}\mathrm{P}_2$ permutation runs. 
Among all models, the Two-Stream Adaptive Graph Convolutional Network (2s-AGCN) and the 1D Convolutional Neural Network (1D CNN) achieved the highest accuracy and F1 score, along with comparatively lower standard deviations, demonstrating strong performance for the intent inference task~(Table~\ref{tab:model_performance}). 

The 2s-AGCN exhibited slightly higher AUC-ROC results, demonstrating superior threshold-invariant ranking capability. STGCN-based 2s-AGCN models remain a strong prospect for action recognition tasks, in part because they incorporate joint detection scores as input, which enhances robustness in noisy data settings. In our implementation—following the approach of Jaiswal and Srivastava~\cite{jaiswal2024benchmarking}—we reduced the original 2s-AGCN architecture from nine to three adaptive graph convolutional network (AGCN) blocks to mitigate overfitting risks associated with smaller sports datasets.

Other models, despite achieving somewhat similar mean scores, degrade on measures of variability and uncertainty with higher standard deviations and wider 95\% confidence intervals, indicating less stable performance. Across all metrics, convolutional and graph-based models consistently outperformed traditional feature-based and sequence models. For subsequent analyses, we selected the 1D CNN model due to its simple architecture, reduced parameter count, and competitive near-real-time performance compared to more complex alternatives. Additionally, the model demonstrates relatively tighter standard deviations and confidence intervals, further highlighting its robust generalizability to unseen data.

\subsection{Performance with Varying Input Length}

To better understand the model's capability to infer batter's intent from temporal pose data, we analyze the classification performance across different input durations (\Cref{tab:threshold_metrics}). As expected, with shorter durations—which likely represent batter's movement before the ball reaches the batter—the model performance is poor, with lower accuracy, AUC-ROC, and F1 scores. On the higher end of the clip-length range, performance plateaus around 80 frames (approximately the dataset’s mean plus one standard deviation length), suggesting diminishing returns beyond this point. Overall, as the duration of the input clip increases, all three performance metrics improve progressively, indicating that longer pose sequences allow the model to more reliably identify motion intent. This result aligns with our labeling approach, which uses bat speed as a proxy for underlying intent, so it is likely that the model requires sufficient temporal context to capture the batter's motion during shot execution in order to infer intent accurately.

Notably, even with input clips of 30 or 40 frames—both shorter than the dataset's mean clip length (55 frames)—the model achieves reasonable accuracy. We take this as an indication that signs of intent are visible even early on during the batter's motion. This finding is promising, as it highlights the value of pose-based analysis for intent inference and supports the design of our temporal modeling approach.

\begin{table}[tbhp]
\centering

\begin{tabular}{@{}crrr@{}}
\toprule
\textbf{Max Clip Length} & \textbf{Accuracy} & \textbf{AUC} & \textbf{F1 Score} \\
\midrule
3   & 0.58 & 0.52 & 0.58 \\
10  & 0.60 & 0.55 & 0.60 \\
20  & 0.64 & 0.65 & 0.65 \\
30  & 0.70 & 0.74 & 0.71 \\
40  & 0.74 & 0.80 & 0.75 \\
50  & 0.76 & 0.82 & 0.76 \\
60  & 0.78 & 0.83 & 0.78 \\
70  & 0.78 & 0.84 & 0.78 \\
80  & 0.78 & 0.84 & 0.79 \\
\bottomrule
\end{tabular}
\caption{
Model performance metrics across different video segment lengths. 
\textit{Pose segment statistics: mean=54.3 frames, median=50.0, mode=46, std=25.7, min=25, max=377.}
}
\label{tab:threshold_metrics}
\end{table}

\showornot{
\textbf{ABLATION STUDY OF FEATURE COMPARISON - SKIPPED-check with range model}
on comparing relevance of different features on classification we find that .... were more important than other which go well with existing understanding of cricket biomechanics.
Feature importance  . Figure  + these features are stable across different dataset and comparisons (Reliable and robust) - they are important in determining energy

\textbf{check how does predictive performance reduce if we reduce input size from models}
}





\begin{table}[tbhp]
    \centering
    
    \vspace{0.5em}
    
    \begin{tabular}{@{}lccc@{}}
        \toprule
        \textbf{Model} & \textbf{Acc} & \textbf{AUC-ROC} & \textbf{F1} \\
        \midrule
        
        LSTM Classifier & 0.68 & 0.72 & 0.64 \\
        LSTM Autoencoder & 0.68 & 0.66 & 0.63 \\
        Motion Range Classifier & 0.67 & 0.72 & 0.57 \\
        2sAGCN & \textbf{0.74}& \textbf{0.81} & \textbf{0.73}\\

        1D CNN         & \textbf{0.74} & 0.76 & \textbf{0.73} \\

        \bottomrule
    \end{tabular}
    \caption{Performance on a single batter's data as test set.}
    \label{tab:kohli_testset}

\end{table}

\section{Case Study for Single Batter Performance}
\label{case_study}
\begin{figure*}[bthp]
    \centering
        \includegraphics[width=0.73\linewidth]{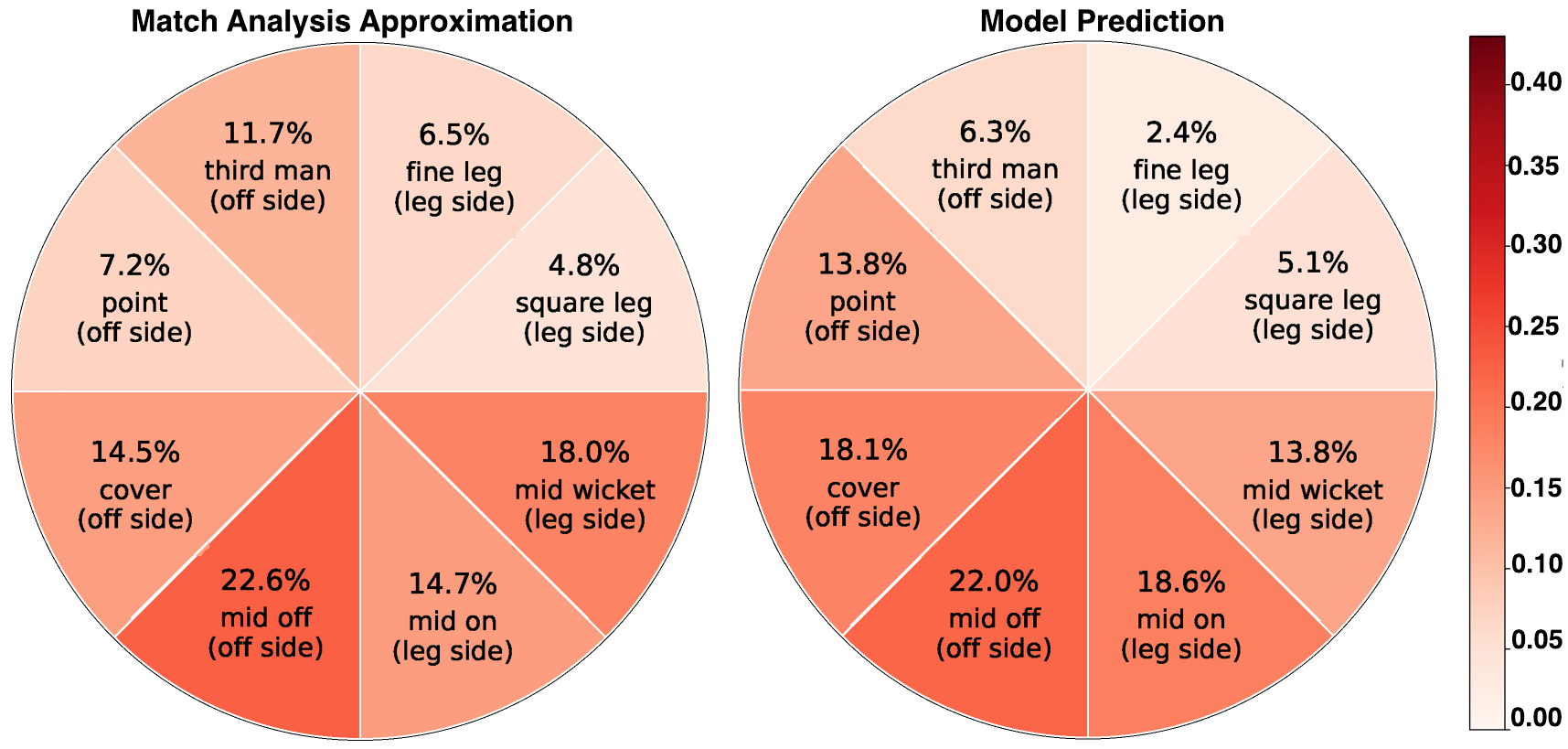}
    \caption{Comparison of high-energy shot region distributions for a single batter over multiple matches: statistics-derived data (35 matches) vs. model prediction (14 matches).}
    \label{fig:kohli_wagon_wheel_high}
\end{figure*}

\begin{figure*}[tbhp]
    \centering
        \includegraphics[width=0.73\linewidth]{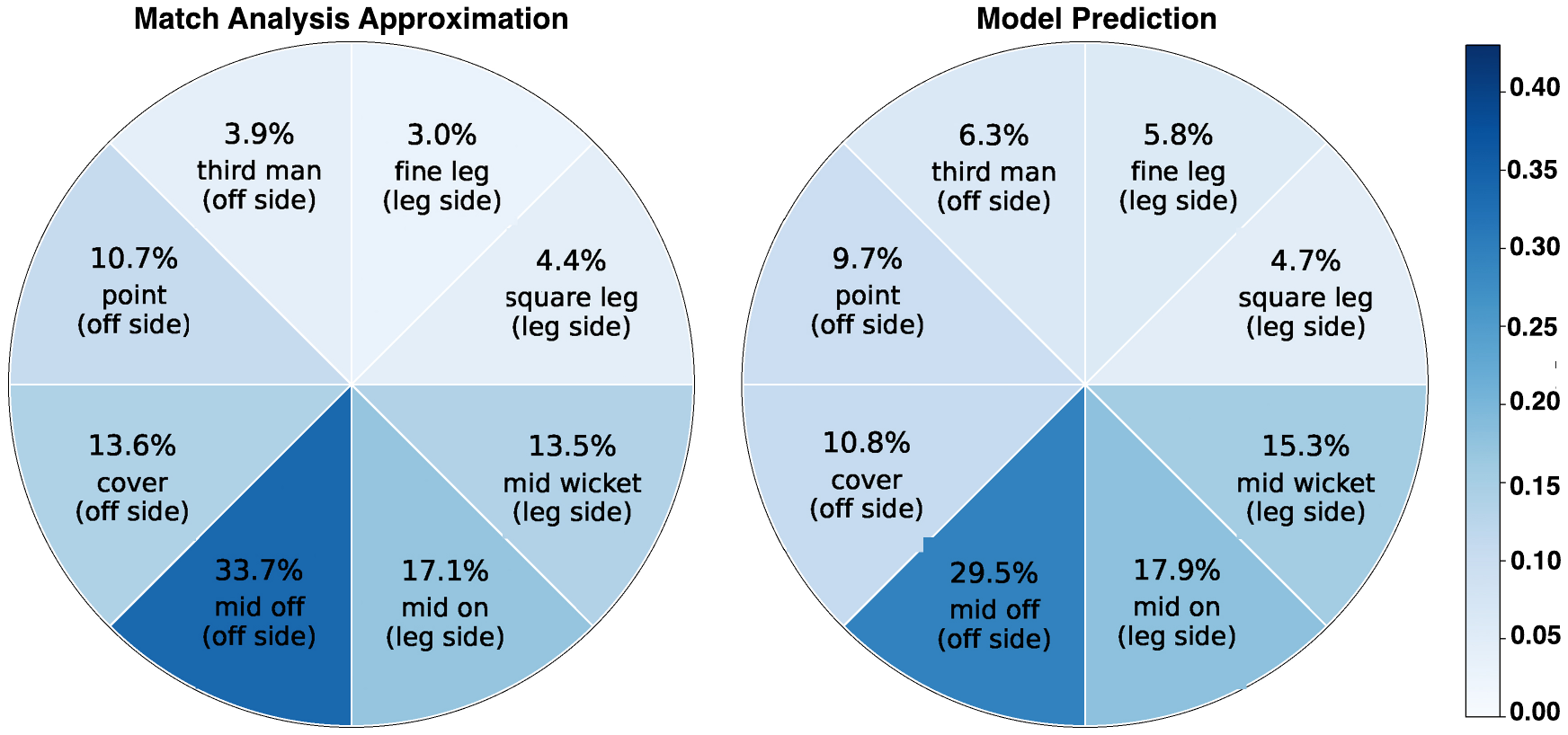}
    \caption{Comparison of low-energy shot region distributions for a single batter over multiple matches: statistics-derived data (35 matches) vs. model prediction (14 matches).}
    \label{fig:kohli_wagon_wheel_low}
\end{figure*}

As there are no established methods to directly validate the correctness of our model’s predictions, we employ several innovative approaches using data from a single batter to assess our results. Specifically, we compare statistics informed by domain knowledge of the batter's cricket shot selection. In cricket, the playing field can be roughly divided into eight regions where shots can be played (Figure~\ref{fig:cric_field}). Each batter, according to their natural style, exhibits preferred (strong) and less favored (weak) regions for shot selection. Analyzing a large sample of games reveals these individualized playing patterns.

To this end, we compute various statistics for one batter using existing match analysis spanning thirty-five matches. In parallel, we curate and annotate a comprehensive video dataset covering fourteen matches for the same batter, and calculate model prediction scores on this data. Table~\ref{tab:kohli_testset} presents the test-time performance of various models on this dataset, which is consistent with our previous results (Table~\ref{tab:model_performance}).

We assess the correspondence between match analysis statistics, our annotated data, and the statistics derived from the 1D CNN model’s predictions. For this evaluation, we compare high- and low-energy shots across the three sources in two ways:
\begin{itemize}
    \item \textbf{Distribution Deviation:} We compare our model's predicted distribution of high- and low-energy shots across all eight field regions against the distribution calculated using match analysis statistics, validating the model's ability to capture the batter's typical shot region preferences for high and low energy shots.

    \item \textbf{Distribution Proportion:} 
    We examine the proportion of high- and low-energy shots within each field region separately, using ground truth annotations from fourteen matches to compare against the model-predicted proportions. Since the number of low-energy shots increases more rapidly with the addition of matches, potentially biasing the proportion estimates, we ensure a fair comparison by using the same fourteen matches for both model-predicted and ground-truth proportions.
\end{itemize}

\subsection{Match Statistics based Distribution Comparisons}

The analysis pipeline—illustrated in Figure~\ref{fig:commentary_pipeline}—outlines the workflow for identifying shot energy and region from both match analysis statistics and model predictions.

The match analysis data provides, for each ball, information about the batter, the runs scored, and the associated shot region. To classify shots from this data as high- or low-energy, we use a simple heuristic: if the runs scored on a ball are $\geq$ 3, the shot is labeled as high-energy; if the runs are $\leq$ 1, the shot is labeled as low-energy. We hypothesize that using data from many games stabilizes the variations in trends of shot area distribution introduced by our approximation technique. 

For the model predictions, each output includes a high- or low-energy classification for every shot. To assign a shot region, we match the ball number (referred to as the over in cricket) from the video data to its corresponding region in the match statistics. This approach enables us to construct the distributions of predicted shots.

The statistical data serves as an approximate ground truth for comparison against model predictions. By determining shot energy using our model's predictions and the existing statistical data, we can compare the shot energy across different regions of the cricket field.

\textbf{Distribution Deviation Results:} Figure~\ref{fig:kohli_wagon_wheel_high} and Figure~\ref{fig:kohli_wagon_wheel_low} visualize the distributions of high- and low-energy shots, respectively, comparing match-derived statistics with model predictions. The figures reveal a high degree of overlap between the model and the approximate statistics, with only minor discrepancies in specific regions. This alignment provides further evidence of our model's effectiveness in inferring shot energy across various areas of the cricket field where the batter plays shots. 

Please see supplementary file Section 1 for distribution deviation results using ground truth labels from fourteen matches.

\begin{table}[tb]
\centering
\setlength{\tabcolsep}{3pt}

\begin{tabular}{@{}lrrrr@{}}
\toprule
\textbf{Side (Total Shots)} & 
\multicolumn{2}{c}{\textbf{High Ratio}} & 
\multicolumn{2}{c}{\textbf{Low Ratio}} \\
\cmidrule(lr){2-3} \cmidrule(lr){4-5}
 & \textbf{True} & \textbf{Model} & \textbf{True} & \textbf{Model} \\
\midrule
cover (143)        & 0.48 & 0.53 & 0.52 & 0.47 \\
fine leg (46)      & 0.33 & 0.22 & 0.67 & 0.78 \\
mid off (273)      & 0.32 & 0.33 & 0.68 & 0.67 \\
mid on (183)       & 0.37 & 0.41 & 0.63 & 0.59 \\
mid wicket (152)   & 0.43 & 0.38 & 0.57 & 0.62 \\
point (117)        & 0.39 & 0.50 & 0.61 & 0.50 \\
square leg (50)    & 0.38 & 0.42 & 0.62 & 0.58 \\
third man (65)     & 0.34 & 0.38 & 0.66 & 0.62 \\
\bottomrule
\end{tabular}
\caption{True Distribution vs Model Prediction Distribution for Proportion Deviation, with total shot counts for the region. All data computed from 14 matches.}
\label{tab:prop_dev}
\end{table}


\textbf{ Proportion Deviations Results:}
Table \ref{tab:prop_dev}  presents a detailed comparison between the ground truth and model-predicted values for proportions of high- and low-energy shots across various fielding positions.  These results indicate that our model’s predictions are well-aligned with the statistical trends for most positions, except for few instances of deviations in the proportion table.

Notably, the model’s predictions for “fine leg” and “point” positions showed greater deviation, which may reflect the inherent variability and difficulty in visual identification, or lower sample sizes at these positions (for fine leg).
These results highlight the importance of robust data collection and the need for careful interpretation when model predictions differ from ground truth in specific contexts. 
Please see supplemental file Section 2 for details on area-wise shot count for the fourteen matches dataset used in this evaluation.

\subsection{Baselines}

Table \ref{tab:baseline_kohli} compares our model’s performance against two baselines—a random predictor and a run-based heuristic—using accuracy, distribution deviation, and average proportion deviation metrics. Across all evaluated metrics, our 1D CNN model most closely aligns with the ground truth statistics. The accuracy score for the 1D CNN differs slightly from those reported in Table \ref{tab:kohli_testset} because some shots were excluded due to unsuccessful over and shot area extraction in the baseline models. To ensure a fair comparison, the same subset of data was used for the 1D CNN predictions.

The random prediction model achieves close to 50\% accuracy, as expected, but exhibits a very high distribution deviation since its random predictions do not account for the specific region of the shot. The heuristic-based approximation model attains slightly better accuracy but much higher deviation scores, since certain areas of the cricket field (like the region behind the batter) allow for higher runs to be scored with less energy, thereby violating the heuristic’s assumption.

These results indicate that visual classification has strong potential to approach human-level judgment, thus providing a robust framework for shot intent inference.



\subsection{One Match Detailed Analysis}
To demonstrate the practical utility of our energy-based shot analysis, we examine a case study from the third One-Day International (ODI) between India and South Africa, played at Cape Town in February 2018. In this match, Indian batter Virat Kohli scored 160 runs off 159 balls against South Africa.

Table \ref{tab:energy_summary} summarizes the distribution of low- and high-energy shots played by Kohli during different phases of his innings (over ranges) using 1D CNN prediction. Our results reveal a progressive shift in energy expenditure: Kohli increased high-energy shots as the innings progressed, particularly accelerating in the final overs. This pattern reflects purposeful energy conservation initially, followed by an aggressive finish—a hallmark of strategic ODI batting.

Table \ref{tab:kohliVsBowler} further breaks down his performance against individual bowlers. We observe that Kohli adopted a more aggressive shot selection against certain bowlers (e.g., Tahir), while opting for energy-conserving, lower-risk shots against others (e.g., Rabada). Such variations reflect context-specific strategy and adaptability to different bowling styles and match situations, effectively captured by our model analysis.

These insights demonstrate the model’s ability to decode context-specific batting strategies, offering granular analytics for training interventions. This case study illustrates how energy-based shot classification complements traditional metrics (such as runs per bowler), enabling deeper tactical insights into a batter’s adaptability, which could be useful both to the batting and the bowling sides.

\begin{table}[tb]
\centering
\setlength{\tabcolsep}{3pt}  
\begin{tabular}{@{}lccc@{}}
\toprule
\textbf{Method} & 
\makecell{\textbf{Accuracy}\\(\%)} & 
\makecell{\textbf{Dist.}\\\textbf{Deviation}} & 
\makecell{\textbf{Avg. Proportion}\\\textbf{Deviation}} \\
\midrule
Random        & 46.9  & 34.90 & 14.9 \\
Runs Approx.  & 66.2  & 28.97 & 22.3 \\
\textbf{1D-CNN} & \textbf{71.4} & \textbf{15.69} & \textbf{5.6} \\
\bottomrule
\end{tabular}
\caption{Model Baseline Comparison against Ground Truth Labels. Distribution deviation calculates the sum of total deviation for high- and low-energy shot distribution. Average proportion deviation refers to the mean deviation within each shot region when comparing high- and low-energy shots. All statistics are based on 14 matches data of a single batter. Total High Shots: 443; Total Low Shots: 679.}
\label{tab:baseline_kohli}
\end{table}

\begin{table}[tb]
\centering

\begin{tabular}{llcc}
\toprule
\textbf{ID} & \textbf{OverRange} & \textbf{Low Energy} & \textbf{High Energy} \\
\midrule
0 & 0--10   & 22 & 6  \\
1 & 10--20  & 11 & 4  \\
2 & 20--30  & 14 & 14 \\
3 & 30--40  & 10 & 13 \\
4 & 40+     & 9  & 17 \\
\bottomrule
\end{tabular}
\caption{Energy Summary: Model Prediction vs. OverRange.}
\label{tab:energy_summary}
\end{table}






\begin{table}[tbhp]
\centering

\begin{tabular}{l c c c c}
\hline
\multirow{2}{*}{\makecell{\textbf{Bowler}}}
& \multirow{2}{*}{\makecell{\textbf{Total}\\\textbf{Runs}}}
& \multicolumn{2}{c}{\textbf{Shot Energy}} 
& \multirow{2}{*}{\makecell{\textbf{Total}\\\textbf{Balls Faced}}} \\
\cline{3-4}
& & \textbf{High} & \textbf{Low} & \\
\hline
Duminy        & 31 & 10 & 18 & 28 \\
Rabada        & 25 & 11 & 19 & 30 \\
Tahir         & 23 & 17 & 5  & 22 \\
Morris        & 22 & 8  & 10 & 18 \\
Ngidi         & 18 & 2  & 12 & 14 \\
Phehlukwayo   & 14 & 6  & 2  & 8  \\
\hline
\end{tabular}
\caption{Total Runs and High/Low Energy Shot Count Against Each Bowler.}
\label{tab:kohliVsBowler}
\end{table}

\section{Conclusion and Discussion}
\label{discussion}

In this work, we focus on mental state inference using video data. To prepare such an intent-driven dataset, we examine sports settings and choose cricket as a representative sport. Based on temporal posture data, we developed a robust framework to infer a batter's intent for shots. Additionally, we constructed a comprehensive dataset of batters' shots with aggressive and defensive intents to validate our hypothesis. Despite inherent noise in posture estimation and the subjective nature of labelling actions as high- and low-energy, we achieved an F1 score exceeding 75\% in distinguishing shot intents.

Our findings have several important implications. First, an automated tool that infers batter's intent from visual cues can assist player training. Coaches and analysts can use such tools to track a batter's weakness,  under stressful situations, in different playing environments, and against particular bowlers. For the bowling side, these insights can help devise strategies to exploit a batter's weak shot regions. More importantly, this approach opens up opportunities for automatic fatigue assessment by tracking a batter's energy expenditure and identifying unusual low-energy shot patterns, which can reduce injury risk, a very critical issue in all sports. By relying solely on postural data, our method opens avenues for monitoring and evaluating athletes across various sports.

Beyond sports, intent inference has promising applications in healthcare and surveillance as well. Using pose and motion, non-invasive vision-based analysis could be explored to identify signals of panic, stress, anger, depression, or violent intent, contributing to more personalised clinical treatment. Vision methods could have a promising role, especially in remote areas with limited access to healthcare facilities, potentially lowering costs and response times. Further research in this area could provide an important direction for advancing non-contact health monitoring.

While our results are promising, it is crucial to acknowledge areas for improvement. Vision-based inference requires more extensive analysis on various datasets to bring them to a more generalizable, practical standard, especially amidst challenges posed by noisy data and subjectivity of intent labels. Future extensions to this study could focus on integrating more sophisticated commentary-derived heuristics as alternative labeling sources, to allow for testing on larger datasets. Similarly, advanced natural language techniques could generate more context-aware labels for intent analysis~\cite{miraoui2023analyzing}.  Extending this approach to other sports and broader populations, with more detailed labelling of mental states, will contribute to its practical impacts.

By leveraging video-based analysis, we present a non-intrusive assistive system to improve response and safety, elucidating broader implications in health and sports settings. The field of human biomechanics offers valuable signals for analytics and clinical applications. This work demonstrated the potential of these biomechanical applications and highlights the pressing need to develop tools that can fully realise their capabilities.

\newpage

\bibliographystyle{unsrtnat}
\bibliography{main}
\end{document}